\documentclass[conference]{IEEEtran}
\IEEEoverridecommandlockouts
\usepackage{cite}
\usepackage{amsmath,amssymb,amsfonts}
\usepackage{algorithmic}
\usepackage{graphicx}
\usepackage{textcomp}
\usepackage{xcolor}
\usepackage{float}
\def\BibTeX{{\rm B\kern-.05em{\sc i\kern-.025em b}\kern-.08em
    T\kern-.1667em\lower.7ex\hbox{E}\kern-.125emX}}

\DeclareMathOperator*{\argmin}{arg\,min}
\DeclareMathOperator*{\argmax}{arg\,max}

\begin{document}

\title{Retrospective motion correction in MRI using disentangled embeddings}

\author{\IEEEauthorblockN{Qi Wang}
\IEEEauthorblockA{
\textit{Department of Radiology}\\
\textit{University of Tuebingen}\\
Tuebingen, Germany \\
wq.levi@gmail.com}
\and
\IEEEauthorblockN{Veronika Ecker}
\IEEEauthorblockA{
\textit{Department of Radiology}\\
\textit{University of Tuebingen}\\
Tuebingen, Germany \\
veronika.ecker@med.uni-tuebingen.de}
\and
\IEEEauthorblockN{Marcel Früh}
\IEEEauthorblockA{\textit{ Department of Radiology}\\
\textit{University of Tuebingen}\\
Tuebingen, Germany \\
marcel.frueh@med.uni-tuebingen.de}
\and
\IEEEauthorblockN{Sergios Gatidis}
\IEEEauthorblockA{\textit{Department of Radiology}\\
\textit{Stanford University}\\
Stanford, California, USA \\
sergios.gatidis@med.uni-tuebingen.de}
\and
\IEEEauthorblockN{Thomas Küstner}
\IEEEauthorblockA{\textit{Department of Radiology}\\
\textit{University of Tuebingen}\\
Tuebingen, Germany \\
thomas.kuestner@med.uni-tuebingen.de}
}
%\documentclass{article}
%\usepackage{spconf,amsmath,graphicx}
%\usepackage{lipsum}
%\usepackage{float}%

%% Example definitions.
%% --------------------
%\def\x{{\mathbf x}}
%\def\L{{\cal L}}%

%\DeclareMathOperator*{\argmin}{arg\,min}
%\DeclareMathOperator*{\argmax}{arg\,max}%

% Title.
% ------
%\title{Retrospective motion correction in MRI using %disentangled embeddings}
%%%
% Single address.
% ---------------
%%\name{Qi Wang, Veronika Ecker, Marcel Früh, Sergios %Gatidis, Thomas Küstner\thanks{Thanks to XYZ agency for %funding.}}%
%\address{Author Affiliation(s)}
%
%% For example:%
% ------------
%\address{School\\
%	Department\\
%	Address}
%
% Two addresses (uncomment and modify for two-address case).
% ----------------------------------------------------------
%\twoauthors
%  {A. Author-one, B. Author-two\sthanks{Thanks to XYZ agency for funding.}}
%	{School A-B\\
%	Department A-B\\
%	Address A-B}
%  {C. Author-three, D. Author-four\sthanks{The fourth author performed the work
%	while at ...}}
%	{School C-D\\
%	Department C-D\\
%	Address C-D}
%
%\name{Qi Wang, Veronika Ecker, Marcel Früh, Sergios Gatidis, Thomas Küstner}
%\address{Medical Image and Data Analysis (MIDAS.lab)\\
%Department of Diagnostic and Interventional Radiology, University of Tuebingen, \\
%Tuebingen, Germany}

%\address{Department of Radiology, Stanford University, Stanford, California, USA
 
%\begin{document}
%\ninept
%
\maketitle

\begin{abstract}
    %Physiological motion remains one of the major extrinsic factors that can impact diagnostic quality of magnetic resonance imaging. While multiple retrospective motion correction methods have been proposed, their effectiveness to generalize for varying motion types and body regions remains limited. In particular, machine learning (ML) driven corrections can be application- and data-specific. We hypothesize although distinct, motion artifacts share a common pattern that could be exploited. In ML, disentangled embeddings have shown to improve the robustness and generalizability of models. In this work, we propose to leverage such merits to recover MR images from simulated motion artifacts by disentangling the motion factor. To this end, a hierarchical vector quantized (VQ) variational auto-encoder is used to encode the motion-to-clean feature continuum into a disentangled codebook. An auto-regressive model is trained to match the prior distribution of the learned codebooks. At inference, the codebook is used for conditional motion correction via an auto-regressive model that matches the latent prior. Meanwhile, the multi-resolution codebooks also capture coarse-to-fine motion features. The disentangled embedding circumvents the need for explicitly training each specific motion artifact. We validate our hypothesis on simulated motion for a whole-body imaging. A strong motion correction robustness on different levels of motion severity was observed. The proposed approach is promising to be used for various motion types and body regions.    
    Physiological motion can affect the diagnostic quality of magnetic resonance imaging (MRI). While various retrospective motion correction methods exist, many struggle to generalize across different motion types and body regions. In particular, machine learning (ML)-based corrections are often tailored to specific applications and datasets. We hypothesize that motion artifacts, though diverse, share underlying patterns that can be disentangled and exploited. To address this, we propose a hierarchical vector-quantized (VQ) variational auto-encoder that learns a disentangled embedding of motion-to-clean image features. A codebook is deployed to capture finite collection of motion patterns at multiple resolutions, enabling coarse-to-fine correction. An auto-regressive model is trained to learn the prior distribution of motion-free images and is used at inference to guide the correction process. Unlike conventional approaches, our method does not require artifact-specific training and can generalize to unseen motion patterns. We demonstrate the approach on simulated whole-body motion artifacts and observe robust correction across varying motion severity. Our results suggest that the model effectively disentangled physical motion of the simulated motion-effective scans, therefore, improving the generalizability of the ML-based MRI motion correction. Our work of disentangling the motion features shed a light on its potential application across anatomical regions and motion types.
    
\end{abstract}

\begin{IEEEkeywords}
    Representation learning, Abdominal MRI, Motion correction
\end{IEEEkeywords}
\section{Introduction}

%background intro
Magnetic resonance imaging (MRI) is widely used in clinical practice for high-resolution imaging of human tissues of different body parts. Despite its advantages, MRI quality is often compromised by physiological and patient movement during the scan. As one of the most common quality impairment, motion artifacts can lead to misinterpretation of images, affecting diagnosis and treatment planning~\cite{moco}. Motion artifacts can occur in different body parts and depending on the type of motion manifest as blurring, ghosting, or misalignment of structures~\cite{moco-complex}. Based on the spatial type of motion - rigid, affine or elastic/non-rigid - and its temporal occurrence - coherent or incoherent - various motion correction methods can be employed to mitigate motion artifacts~\cite{moco, rigid_moco}. While prospective approaches aim to acquire motion-free data by for example faster imaging~\cite{prosepctive_moco}, triggered acquisition~\cite{prosepctive_moco, retro_ext}, motion-corrected gradients~\cite{nagivator, navigator_rapid, vaillant, grappa}, retrospective methods rely on recovering the motion-corrected image from motion-corrupted data~\cite{moco_prior, moco_cnn}, . Data-driven motion correction methods~\cite{medgan, moco_cnn, moco_prior} aim to address this problem by translating the motion-corrupted image to their clean counterparts using generative models. These models hence learn a mapping from motion-corrupted images to clean images, effectively removing artifacts and by that improving image quality ~\cite{moco_survey}.

%specific problem
Most of these data-driven approaches use a generative model formulated as image-to-image task, with convolutional networks for identification of the motion aliasing patterns. Previous works hereby operated either on image patches sampled at multiple resolutions, and which were passed to a convolution network to capture the features of motion artifacts~\cite{moco_cnn}. Similarly, convolution auto-encoder was employed to extract the representative motion artifact features for the reconstruction of the motion-free image from the latent space embedding~\cite{moco_prior, lee2021, ksz2020}. Commonly, Unet architectures~\cite{unet} serve as the convolutional network backbone whose skip-connection can help to preserve detailed spatial image information at different resolutions~\cite{moco_prior,ksz2020, pawar2020, almasni2021}. The capability of generative adversarial networks (GANs)~\cite{gan} enabled more realistic motion-free image generation by employing perceptual loss~\cite{medgan, mocogan}. Conditional GAN generates motion-free images, which are supported by its discriminator, from motion-corrupted images~\cite{cgan_moco}. Similarly, image reconstruction from k-space is incorporated to the motion-correction task during the forward pass in a conditional GAN~\cite{mutlishot_gan_moco}.

%knowledge gap
Current data-driven methods leverage multi-resolution of the motion features, and achieve perceptually realistic motion-free images. They formulate the image generation and motion-correction as one task~\cite{medgan, mocogan, moco_cnn, moco_prior, cgan_moco, mutlishot_gan_moco}. Despite these advancements, the variety of motion patterns (spatial and temporal) as well as the motion occurence and appearance in different body regions remain difficult to capture and correct for a single method. Moreover, insufficient training samples usually exist for training a generalizable method. These limitations thus hinder their adoption in real-world scenarios where models are tasked to perform on a range of motion types. We hypothesize that although apparently different in image appearance, motion features among body regions share common features and that confounding (body regions, type and severity of motion) can be disentangled to different dimensions, to ultimately allow for a more comprehensive representation of the motion characteristics. In this work, we explore a first step towards this goal. We propose to train a generative model that disentangles the severity of motion artifacts on the perceived image quality. In contrast to the pre-dominant models in this field (GAN, DDPM), we propose a vector quantized (VQ) model to capture the distribution of motion features by learning a dynamic prior over the latent space~\cite{vqvae}. 
The dynamic prior distribution disentangles motion severity by conditioning on surrogate severity labels, which further entitles the high quality sampling of the latent space~\cite{vqvae2, vqgan}. Our model is trained on paired dataset of motion-free and simulated motion-corrupted data. Unlike previous works~\cite{mocogan, cgan_moco} on image translation tasks, our model learns a conditional prior over the source and target image patterns, which shows a strong robustness over varied motion artifacts and can be extended to incorporate richer context as conditional information.

\section{Methods}
Our proposed method is based on the vector quantized variational autoencoder (VQVAE) architecture, which comprises a two-stage training process. The first stage is to train a VQVAE model on a large dataset of motion-corrupted images, while the second stage is to train another model to match the prior distribution of the learned codebooks. 

\begin{figure}
\centering
    \includegraphics[width=.5\textwidth]{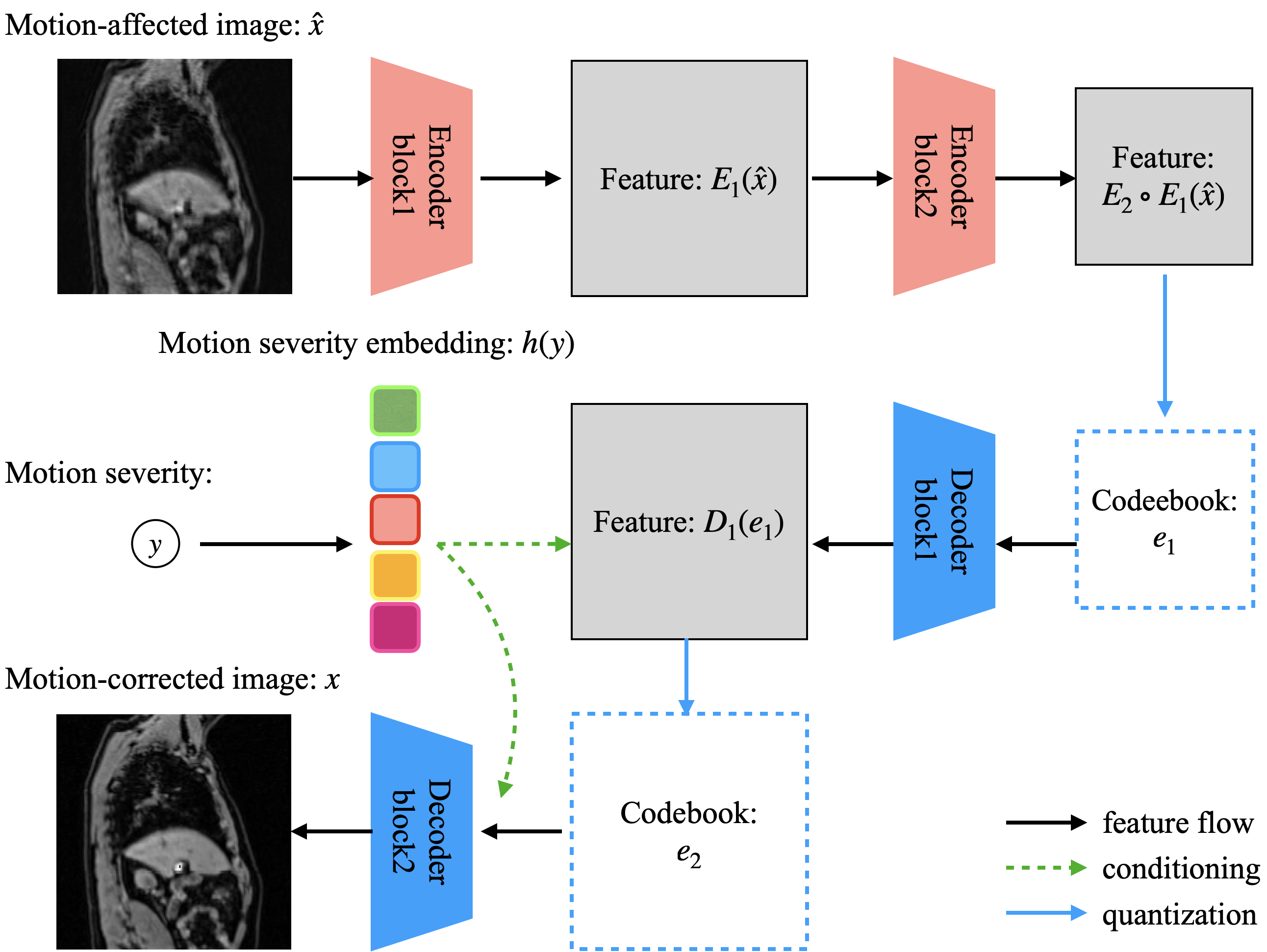}
    \caption{Diagram of our conditional auto-encoder during codebook training. The model takes as input a motion-affected image $\hat{x}$ and performs multi-resolution encoding to extract discrete features $e_{1}$ and $e_{2}$. The class conditioning $y$ is passed through a linear layer $h(\cdot)$ and used to condition on both the feature map $D_{1}(e_{1})$ and codebook $e_{2}$ of higher resolutions to generate a motion-corrected clean target image $x$ through the decoding architecture. The conditioning operation hereby refers to feature add-ups.}
    \label{fig:CVQVAE_y1}
\end{figure}

During the initial training stage, the model comprises an encoder, a decoder, and codebooks of discrete latent vectors at different resolutions. The motion-affected image $\hat{x}$ is input to the encoder of two sequential blocks $E_{1}(\cdot)$ and $E_{2}(\cdot)$, which generate feature activations at multiple resolutions. These activation maps are then quantized to discrete codebooks $e=\{e_{1}, e_{2}\}$, at corresponding resolutions and subsequently decoded through decoder blocks $D_{1}(\cdot)$ and $D_{2}(\cdot)$ to recover motion-corrected images $x$ conditioned on text embeddings of motion severity $h(y)$, as shown in Fig.~\ref{fig:CVQVAE_y1}. The training process involves minimizing the $\ell_1$ distance between the motion-corrected output $x$ and a reference motion-free image $x_\text{ref}$, a codebook loss of Euclidean distance between latent feature and codebook vector $e$, and a commitment loss~\cite{vqvae2} that encourages the encoder to output features which minimize the euclidean distance to codebook vectors. 
Once the autoencoder is trained, in the second training stage, the learning of an auto-regressive model commences to match the prior distribution by predicting the next codebook vector given the previous ones with conditioning on the class vectors of the input image. This allows the model to sample from the learned prior space and generate new images with conditional information on the codebooks and to decode them into the motion-corrected images. 

 \subsection{Vector Quantized Latent Embeddings}
Similar to VQVAE~\cite{vqvae}, the encoding process for latent codebooks, namely the latent feature embedding of the training dataset, uses an encoder-decoder architecture. The codebook is a fixed-size lookup table of reusable feature representations, queried by indices at inference time~\cite{vqvae}. It updates the discrete latent space by minimizing the Euclidean distance of encoder representation for paired inputs of a motion-affected image $\hat{x}$ and a class label $y$. The class label $y$ is defined as an integer parameter for the scope of motion simulation, in our specific case, and quantifies the range of translational motion in $mm$. Specifically, a linear embedding layer $h(\cdot)$ projects the class label $y$ into a class feature vector $h(y)$, which is added to the input image features during the forward pass to enable class conditioning. An $\ell_1$ mean absolute error loss is used to enforce sharper image edges in comparison to the commonly used mean squared error.

During back-propagation, the gradient does not directly pass through the latent codebooks, due to its discrete nature. Instead, a stop-gradient ($\text{sg}$) operation is applied to the output activation from the encoder ($E$) and the codebook vectors ($e$) to update one another. The training loss of the first training stage is given as:

\begin{align}
    \mathcal{L}_\text{first} = \underbrace{\lVert x - x_\text{ref} \rVert_{1}}_{\text{reconstruction loss}}
    + \underbrace{\lVert \text{sg}[E(\hat{x})] - e \rVert^{2}_{2}}_{\text{codebook loss}}
    + \beta \underbrace{\lVert \text{sg}[e] - E(\hat{x})\rVert ^{2}_{2}}_{\text{commitment loss}}
\end{align}
where $\beta=0.25$ is a hyperparameter that controls the weight of the commitment loss for the latent codebooks.

\subsection{Disentangled codebooks}
Commonly, codebooks are used to learn image features from the input images, from which the model tries to approximate a perfect reconstruction, i.e. a motion-corrected image in our work. In our approach, explicit conditioning $y$ is used to disentangle the code vectors $e$ from the input images $\hat{x}$ (see Eq.~\ref{eq:quantization}). In other words, features of different motion severity are disentangled. Specifically, the codebook vector $e_{i}$ is pushed close to the feature vector $\hat{z}$ generated by conditional encoding $E(\hat{x}, y)$. The updated feature representation $z_{q}$ is then input to the decoder for reconstruction.

\begin{figure*}[t]
\centering
\includegraphics[width=.75\textwidth]{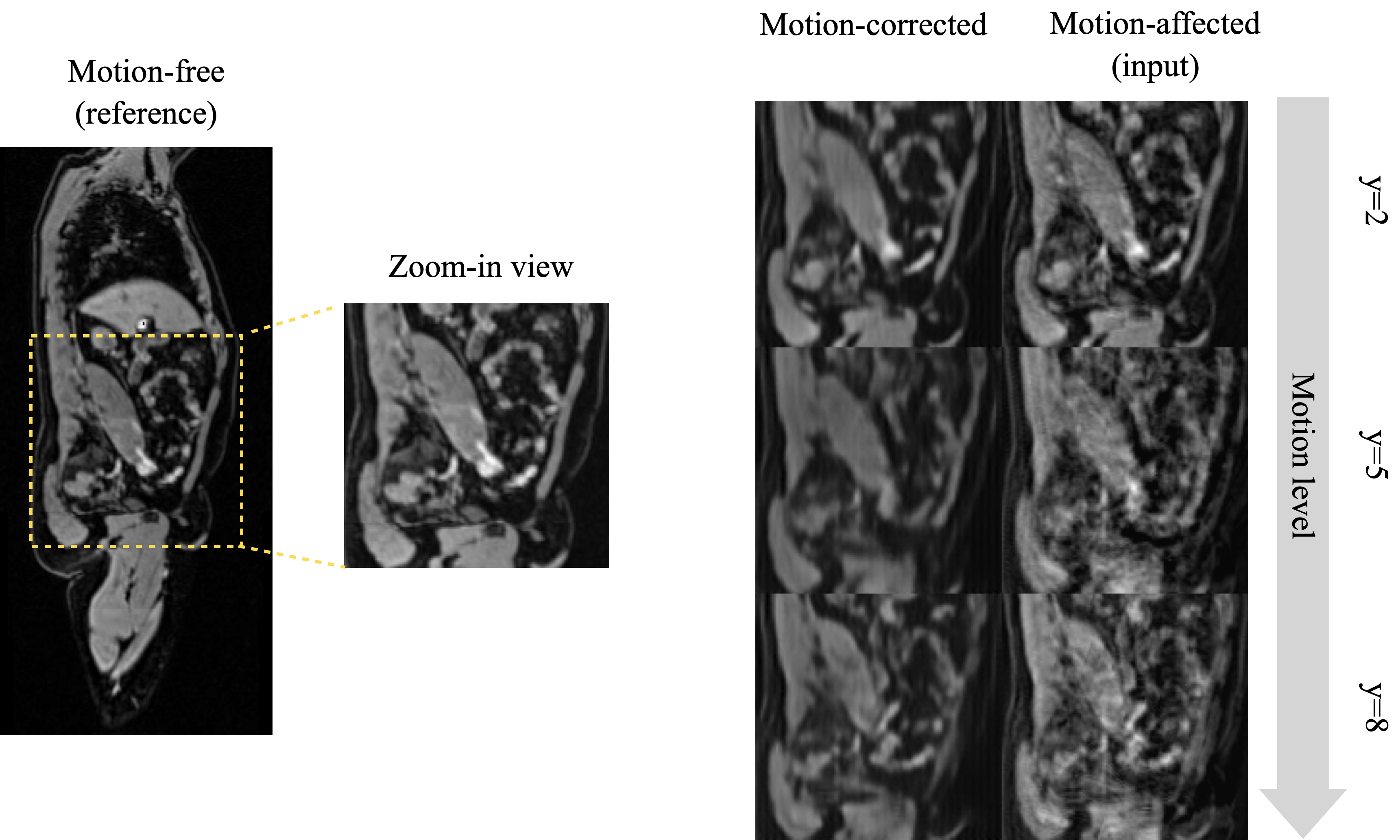}
\caption{Qualitative results of the proposed model for motion correction at various motion severity levels $y$. From the sagittal whole-body imaging plane (left panel), the abdominal region was selected (yellow dashed box), serving as the motion-free reference. The right panel shows the zoomed-in abdominal views with increasing motion severity level (from top to bottom rows), in the motion-corrected (left) and motion-affected (right) images. The model successfully removes motion artifacts while preserving anatomical structures across different motion severity levels.}
\label{fig:CVQVAE_y2}
\end{figure*}

\begin{align}
    z_{q} = \argmin_{e_{i} \in E} ||\hat{z} - e_{i}||_{2}^{2}, \quad \hat{z} = E(\hat{x}, y)
    \label{eq:quantization}
\end{align}
This does not require additional training on difference data source, for example, training separate codebooks for both motion-corrupted and clean dataset to learn a mapping between them. 

%We evaluate the proposed disentangled auto-encoder model on a dataset of motion-corrupted images, at different motion levels (as in Fig.~\ref{fig:CVQVAE_y2}). The results demonstrate that the model successfully cleaned motion artifacts while maintaining consistency with class labels, as shown in Figure~\ref{fig:CVQVAE_y2}. The model effectively disentangles the codebook vectors from the input images, allowing for better generalization to unseen data.

\subsection{Conditional Generation}
The generation process is conditioned on the class label of the input image, which can be represented as a sequence of indices on the learned codebook vectors. To generate new or conditioned images, we use a conditioned auto-regressive model $\theta$ that predicts the next codebook vector given the previous ones: $p(s_{i},c)=\prod_{i}p(s_{<{i}}, c)$. The optimal model $\theta^{*}$ is trained to maximize the likelihood for the sequence of codebook vectors conditioned on the class label:

\begin{align}
    \theta^{*} = \argmax_{\theta} \prod_{i=1}^{N} p(s_{i} | c_{i})  
    \label{eq:auto_regressive}
\end{align}

For the conditional prior matching, the codebook indices of the input images can be sampled with the class label as an additional information. The generated motion-corrected images are thus consistent with the motion level $y$. The generated codebook indices can be decoded into clean images using the decoder of the VQVAE model, which reconstructs images from the sampled codebook vectors.

\subsection{Database}
The training dataset contains 100 quality-controlled 3D volumes of abdominal MRI scans (dual-echo gradient echo imaging, resolution= $2.23\times 2.23\times 3 \text{mm}^3$ to $2.23\times 2.23\times 4.5 \text{mm}^3$) and is a subset of the UKBiobank~\cite{ukb}. The test set contains 30 3D volumes of abdominal scans from the same data repository as the training dataset. These images were screened and are free of motion artifacts. For training and testing, the model takes random crops in the 2D axial slices with matrix size of $128\times 128$. To validate our proposed approach, we used simulated motion. To this end, we applied random affine transformations to the motion-free images, using TorchIO~\cite{torchio}. The class labels $y$ are derived from the amplitude of the motion parameters in the affine transformation. The class labels range from 0 to 10, where 0 indicates no motion and 10 indicates severe motion artifacts. For motion amplitude values between 0 degree of rotation and 0 mm translation to 10 degree of rotation and 10 mm translation along all three axis, a class label of 0 to 10 was assigned.  

\subsection{Model architecture}
Our model is trained in two training phases. The first phase mainly updates parameters in the encoder and decoder. Specifically, the encoder architecture amounts to 2 residual blocks, each hidden channel equals 128; the decoder architecture comprises also 2 residual blocks, each followed by a transpose convolution layer for feature upsampling. The entire encoder-decoder architecture amounts to around 1.13 million trainable parameters. In the second training phase, an autoregressive (AR) model, PixelSNAIL~\cite{pixelsnail}, is implemented to learn the prior distribution over latent space by predicting the next codebook indices from the previous ones. In our setting, the AR model slides across the codebook from top-left corner to the bottom-right. The PixelSNAIL is made up of 4 sub-blocks, with each containing 2 gated ResNet blocks and a causal attention block (details in~\cite{pixelsnail}).

For training, we use an Adam optimizer with default $\beta=(0.9, 0.999)$ and learning rates of $1e^{-4}$ for the first phase and $3e^{-4}$ for the second phase. The complete training for the both phases takes 100 epochs, on an NVIDIA V100 (32GB) GPU.

\section{Results}

%\subsection{Motion correction from the reconstruction}

%\begin{figure*}[t]
%\includegraphics[width=\textwidth]{./figs/figure2.png}
%\label{fig:CVQVAE_y2}
%\caption{Qualitative results of the model on multi-level motion correction. The left panel indicates the entire sagittal slice of the clean target image, where the yellow dashed box is zoomed in as shown in the right panel. The right panel shows zoom-in view of the region marked with yellow dashed box in the left panel. From top to bottom rows are different levels of motion severity used for testing. From left to right columns are the clean target image, the reconstructed image from the model, and the motion-corrupted input image. The model successfully removes motion artifacts while preserving anatomical structures and class labels.}
%\end{figure*}
As we are interested to achieve a perceptual improved image quality, that cannot be easily and reliably captured by any conventional metric, we did refrain from performing any quantitative analysis. We evaluated the qualitative performance of the proposed disentangled auto-encoder model on the test set. The motion-corrupted images were simulated at different motion levels: $y=2$, $y=5$, and $y=8$ (Fig.~\ref{fig:CVQVAE_y2}). The model takes as input the motion-corrupted images and the class labels/motion levels as complementary information, and performs the multi-resolution feature encoding to the discrete latent vectors, which are then decoded to reconstruct a motion-corrected image.

The results demonstrate that the model successfully removes motion artifacts while maintaining consistency across motion severity levels (Fig.~\ref{fig:CVQVAE_y2}). For mid ($y=5$) to strong ($y=8$) motion levels, major Nyquist ghosting patterns are removed, and diffuse motion patterns are deblurred. This suggests that our model has effectively learned to disentangle the codebook vectors from the input images, as well as the successful conditioning on the class labels. We expect this to allow a better generalization to broader levels of motion scenarios. It should be noted that the proposed decoder architecture did not fully recover sharp image edges, especially in the more severe motion-affected cases. 

%\subsection{Motion correction using the conditioned prior}
We used the prior model conditioned on the class label to evaluate how effective the embedding was disentangled. In contrast to the experiment of Fig.~\ref{fig:CVQVAE_y2}, the prior model rearranged the conditioned prior to predict codebook indices of feature representations, instead of directly querying the indices from the encoder's representation.
%Similar to the previous experiment, the model takes input the motion-corrupted images with noise levels as conditioning, and maps to the discrete latent vectors. The auto-regressive model then samples the codebook indices conditioned on noise labels from the learned prior distribution, which are then decoded to reconstruct the clean target images.

\begin{figure}[h]
\centering
\includegraphics[width=.45\textwidth]{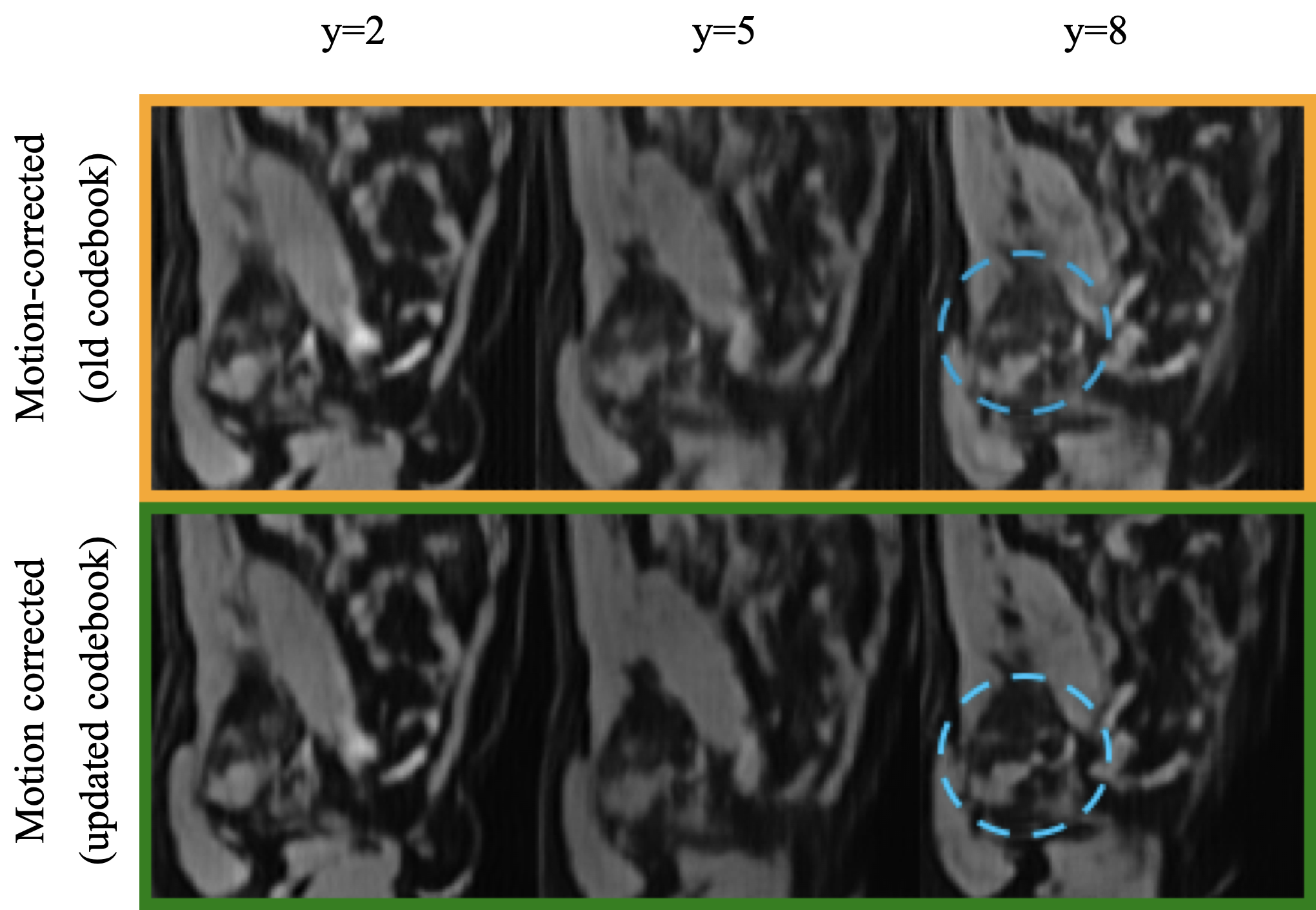}
\caption{Qualitative comparison of the motion-corrected results before and after codebook rearranged with conditioning on prior. The top row shows the motion-corrected results without the codebooks rearranged, encircled in yellow box, while the bottom row shows the motion-corrected results decoded from rearranged codebooks, in green box. Motion severity is increased from left to right, being $y=2, y=5$, and $y=8$ respectively. Note that the low frequency motion patterns are further removed in the lower row, presenting cleaner image content.}
\label{fig:CVQVAE_y3}
\end{figure}

The motion-corrected images show sharper edges and better anatomical preservation (Fig.~\ref{fig:CVQVAE_y3}), as also highlighted by the blue circle. Especially at high motion severity level, localized motion artifacts are better removed (e.g. at $y=5$ and $y=8$).

We acknowledge several limitations of our work, including the less-sharpened image edges and its settings on simulated motion-corrupted images. Moreover, the second training stage operates in a sequential manner and is therefore computationally slow.

\section{Conclusion}
In this work, we proposed a disentangled auto-encoder model for motion artifact correction in MRI images. Our approach leverages a two-stage training process, where the first stage involves training a hierarchical conditional VQVAE model on a paired dataset of motion-corrupted images with motion serverity levels, and motion-free images. In the second stage the model focuses on matching the prior distribution of codebooks conditioned by the motion severity labels using an auto-regressive model.

The overall results from both stages demonstrate the effectiveness of our design to condition on the latent representations, for the task of motion correction while preserving anatomical structures. Our results indicate the benefit of disentangled latent space for image-to-image translation tasks. Future work are expected to explore the application to combine with multi-modality models to enrich the training dataset for various artifacts and the incorporation of additional prior knowledge for improved performance.  
\bibliographystyle{IEEEbib}
\bibliography{refs}

\end{document}